\documentclass[11pt,a4paper]{article}

\usepackage{arabtex} 
\usepackage{utf8}
\usepackage{acl2020}

\usepackage{times}
\usepackage{latexsym}
\usepackage{amsmath}
\usepackage{color}
\usepackage{booktabs}
\usepackage{amsfonts}
\usepackage{graphicx}
\usepackage{subcaption}

\usepackage{tabularx}
\usepackage{inconsolata}
\usepackage{microtype}
\usepackage{CJKutf8}
\usepackage{url}
\usepackage{float}

\aclfinalcopy 
\def\aclpaperid{3218}

\newcommand{\rel}[1]{\texttt{\textbf{#1}}}
\newcommand{\bfsc}[1]{\textbf{\textsc{#1}}}
\newcommand{\bfit}[1]{\textbf{\textit{#1}}}
\newcommand{\h}{\mathbf{h}}

\newcommand{\figref}[1]{Figure~\ref{fig:#1}}
\newcommand{\tabref}[1]{Table~\ref{tab:#1}}

\title{Finding Universal Grammatical Relations in Multilingual BERT}

\author{Ethan A. Chi, John Hewitt, {\normalfont and} Christopher D. Manning \\
Department of Computer Science \\
Stanford University \\
\texttt{\{ethanchi,johnhew,manning\}@cs.stanford.edu}
}

\date{}

\begin{document}

\maketitle

\begin{abstract}
Recent work has found evidence that Multilingual BERT (mBERT), a transformer-based multilingual masked language model, is capable of zero-shot cross-lingual transfer, suggesting that some aspects of its representations are shared cross-lingually.
To better understand this overlap, we extend recent work on finding syntactic trees in neural networks' internal representations to the multilingual setting.
We show that subspaces of mBERT representations recover syntactic tree distances in languages other than English, and that these subspaces are approximately shared across languages.
Motivated by these results, we present an unsupervised analysis method that provides evidence mBERT learns representations of syntactic dependency labels, in the form of clusters which largely agree with the Universal Dependencies taxonomy.  
This evidence suggests that even without explicit supervision, multilingual masked language models learn certain linguistic universals.
\end{abstract}

\section{Introduction}

Past work \cite{liu2019linguistic,tenney2019bert,tenney2019what} has found that masked language models such as BERT \cite{devlin2019bert} learn a surprising amount of linguistic structure, despite a lack of direct linguistic supervision.
Recently, large multilingual masked language models such as Multilingual BERT (mBERT) and XLM \cite{conneau2019cross,conneau2019unsupervised} have shown strong \textit{cross-lingual} performance on tasks like XNLI \cite{lample2019cross,williams2018broad} and dependency parsing \cite{wu2019beto}.
Much previous analysis has been motivated by a desire to explain why BERT-like models perform so well on downstream applications in the monolingual setting, which begs the question: what properties of these models make them so cross-lingually effective?

\begin{figure}[t]
    \centering
    \includegraphics[width=0.95\linewidth]{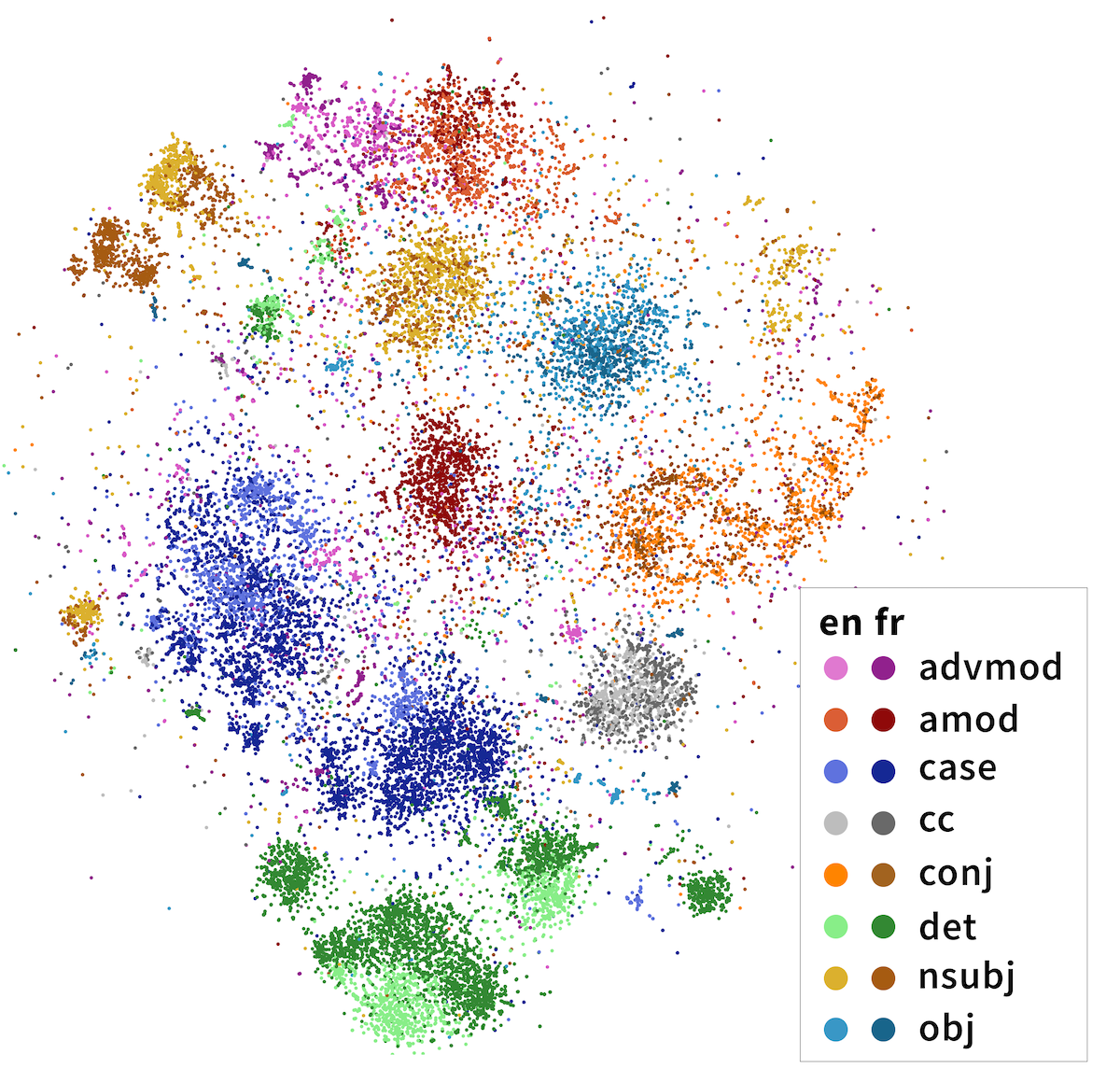}
    \caption{\label{fig:enfr_clusters}t-SNE visualization of head-dependent dependency pairs belonging to selected dependencies in English and French, projected into a syntactic subspace of Multilingual BERT, as learned on English syntax trees. Colors correspond to gold UD dependency type labels. Although neither mBERT nor our probe was ever trained on UD dependency labels, English and French dependencies exhibit cross-lingual clustering that largely agrees with UD dependency labels.} 
\end{figure}

In this paper, we examine the extent to which Multilingual BERT learns a cross-lingual representation of syntactic structure. 
We extend probing methodology, in which a simple supervised model is used to predict linguistic properties from a model's representations.
In a key departure from past work, we not only evaluate a probe's performance (on recreating dependency tree structure), but also use the probe as a window into understanding aspects of the representation that the probe was not trained on (i.e. dependency labels; Figure~\ref{fig:enfr_clusters}).
In particular, we use the \textit{structural probing} method of \citet{hewitt2019structural}, which probes for syntactic trees by finding a linear transformation under which two words' distance in their dependency parse is approximated by the squared distance between their model representation vectors under a linear transformation.
After evaluating whether such transformations recover syntactic tree distances across languages in mBERT, we turn to analyzing the transformed vector representations themselves.

\begin{figure}
    \centering
    \includegraphics[width=0.6\columnwidth]{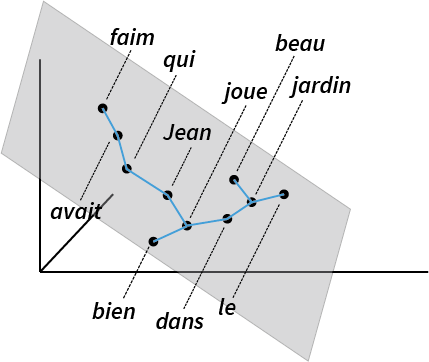}
    \caption{\label{fig:structuralprobe}The structural probe recovers syntax by finding a syntactic subspace in which all syntactic trees' distances are approximately encoded as squared $L_2$ distance \citep{hewitt2019structural}.}
\end{figure}

We interpret the linear transformation of the structural probe as defining a \textit{syntactic subspace} (\figref{structuralprobe}), which intuitively may focus on syntactic aspects of the mBERT representations.
Since the subspace is optimized to recreate syntactic tree distances, it has no supervision about edge labels (such as \textit{adjectival modifier} or \textit{noun subject}).
This allows us to unsupervisedly analyze how representations of head-dependent pairs in syntactic trees cluster and qualitatively discuss how these clusters relate to linguistic notions of grammatical relations.

We make the following contributions:
\begin{itemize}
  \setlength\itemsep{0em}
    \item We find that structural probes extract considerably more syntax from mBERT than baselines in 10 languages, extending the structural probe result to a multilingual setting.
    \item We demonstrate that mBERT represents some syntactic features in syntactic subspaces that overlap between languages. We find that structural probes trained on one language can recover syntax in other languages (zero-shot), demonstrating that the syntactic subspace found for each language picks up on features that BERT uses across languages.
    \item Representing a dependency by the difference of the head and dependent vectors in the syntactic space, we show that mBERT represents dependency clusters that largely overlap with the dependency taxonomy of Universal Dependencies (UD) \cite{nivre2020universal}; see \figref{enfr_clusters}.
    Our method allows for fine-grained analysis of the distinctions made by mBERT that disagree with UD, one way of moving past probing's limitation of detecting only linguistic properties we have training data for rather than properties inherent to the model.

\end{itemize}
Our analysis sheds light on the cross-lingual properties of Multilingual BERT, through both zero-shot cross-lingual structural probe experiments and novel unsupervised dependency label discovery experiments which treat the probe's syntactic subspace as an object of study.
We find evidence that mBERT induces universal grammatical relations without any explicit supervision, which largely agree with the dependency labels of Universal Dependencies.\footnote{Code for reproducing our experiments is available here: \url{https://github.com/ethanachi/multilingual-probing-visualization}}

\section{Methodology}

We present a brief overview of \citet{hewitt2019structural}'s structural probe, closely following their derivation.
The method represents each dependency tree $T$ as a distance metric where the distance between two words $d_T(w_i,w_j)$ is the number of edges in the path between them in $T$.
It attempts to find a single linear transformation of the model's word representation vector space under which squared distance recreates tree distance in any sentence. 
Formally, let $\mathbf{h}^\ell_{1:n}$ be a sequence of $n$ representations produced by a model from a sequence of $n$ words $w^\ell_{1:n}$ composing sentence $\ell$.  
Given a matrix $B \in \mathbb{R}^{k \times m}$ which specifies the probe parameters, we define a squared distance metric $d_B$ as the squared $L_2$ distance after transformation by $B$:
\begin{align*}
d_B(\mathbf{h}^\ell_i, \mathbf{h}^\ell_j) = ||B \mathbf{h}^\ell_i - B \mathbf{h}^\ell_j||^2_2
\end{align*}
We optimize to find a $B$ that recreates the tree distance $d_{T^\ell}$ between all pairs of words ($w_i^\ell$, $w_j^\ell$) in all sentences $s^\ell$ in the training set of a parsed corpus.
Specifically, we optimize by gradient descent: 
\begin{align*}
\arg \min_{B} \sum_{\ell} \frac{1}{|s^\ell|^2} \sum_{i, j} |d_{T^\ell} (w_i^\ell, w_j^\ell) - d_B (\h_i^\ell, \h_j^\ell)|
\end{align*}
For more details, see \citet{hewitt2019structural}.

Departing from prior work, we view the probe-transformed word vectors $B\h$ themselves---not just the distances between them---as objects of study.
The rows of $B$ are a basis that defines a subspace of $\mathbb{R}^m$, which we call the \textit{syntactic subspace}, and may focus only on parts of the original BERT representations. A vector $B\h$ corresponds to a point in that space; the value of each dimension equals the dot product of $\h$ with one of the basis vectors.\footnote{For ease of notation, we will discuss vectors $B\h$ as being in the syntactic subspace, despite being in $\mathbb{R}^k$.}

\subsection{Experimental Settings}

These settings apply to all experiments using the structural probe throughout this paper.

\paragraph{Data}
Multilingual BERT is pretrained on corpora in 104 languages; however, we probe the performance of the model in 11 languages (Arabic, Chinese, Czech, English, Farsi, Finnish, French, German, Indonesian, Latvian, and Spanish).\footnote{When we refer to \textit{all languages}, we refer to all languages in this set, not all languages that mBERT trains on.}$^{,}$\footnote{This list is not typologically representative of all human languages. However, we are constrained by the languages for which both large UD datasets and mBERT's pretraining are available.
Nevertheless, we try to achieve a reasonable spread over language families, while also having some pairs of close languages for comparison. 
}
Specifically, we probe the model on trees encoded in the Universal Dependencies v2 formalism \cite{nivre2020universal}.

\paragraph{Model} In all our experiments, we investigate the 110M-parameter pre-trained weights of the BERT-Base, Multilingual Cased model.\footnote{\url{https://github.com/google-research/bert}}

\paragraph{Baselines} We use the following baselines:\footnote{We omit a baseline that uses uncontextualized word embeddings because \citet{hewitt2019structural} found it to be a weak baseline compared to the two we use.}

\begin{itemize}
\item \textbf{\textsc{mBERTRand}}: A model with the same parametrization as mBERT but no training.  Specifically, all of the contextual attention layers are reinitialized from a normal distribution with the same mean and variance as the original parameters.  However, the subword embeddings and positional encoding layers remain unchanged.  
As randomly initialized ELMo layers are a surprisingly competitive baseline for syntactic parsing \cite{conneau-etal-2018-cram}, we also expect this to be the case for BERT\@.  In our experiments, we find that this baseline performs approximately equally across layers, so we draw always from Layer~7. 
\item \textbf{\textsc{Linear}}: 
All sentences are given an exclusively left-to-right chain dependency analysis.
\end{itemize}

\paragraph{\textsc{Evaluation}} To evaluate transfer accuracy, we use both of the evaluation metrics of \citet{hewitt2019structural}.
That is, we report the Spearman correlation between predicted and true word pair distances (DSpr.).\footnote{Following \citet{hewitt2019structural}, we evaluate only sentences of lengths 5 to 50, first average correlations for word pairs in sentences of a specific length, and then average across sentence lengths.}
We also construct an undirected minimum spanning tree from said distances, and evaluate this tree on undirected, unlabeled attachment score (UUAS), the percentage of undirected edges placed correctly when compared to the gold tree.

\section{Does mBERT Build a Syntactic Subspace for Each Language?}

We first investigate whether mBERT builds syntactic subspaces, potentially private to each language, for a subset of the languages it was trained on;
this is a prerequisite for the existence of a \textit{shared}, cross-lingual syntactic subspace.

Specifically, we train the structural probe to recover tree distances in each of our eleven languages. 
We experiment with training syntactic probes of various ranks, as well as on embeddings from all 12 layers of mBERT.

\begin{table*}
\small
\centering
\begin{tabular}{lrrrrrrrrrrrrr}
\multicolumn{13}{c}{\textbf{Structural Probe Results: Undirected Unlabeled Attachment Score (UUAS)}}\\
\toprule
                    &\tiny \bf Arabic    &\tiny \bf Czech    &\tiny \bf German    &\tiny \bf English    &\tiny \bf Spanish    &\tiny \bf Farsi    &\tiny \bf Finnish    &\tiny \bf French    &\tiny \bf Indonesian    &\tiny \bf Latvian    &\tiny \bf Chinese    &\tiny \bf Average    \\
\bfsc{Linear} & 57.1 & 45.4 & 42.8 & 41.5 & 44.6 & 52.6 & 50.1 & 46.4 & 55.2 & 47.0 & 44.2 & 47.9 \\
\bfsc{mBERTRand} & 49.8 & 57.3 & 55.2 & 57.4 & 55.3 & 43.2 & 54.9 & 61.2 & 53.2 & 53.0 & 41.1 & 52.9 \\
\midrule
\bfsc{In-Lang} & 72.8 & 83.7 & 83.4 & 80.1 & 79.4 & 70.7 & 76.3 & 81.3 & 74.4 & 77.1 & 66.3 & 76.8 \\
$\Delta_{\text{\bfsc{Baseline}}}$ & 15.7 & 26.4 & 28.1 & 22.6 & 24.1 & 18.0 & 21.4 & 20.1 & 19.1 & 24.1 & 22.1 & 22.0 \\
\midrule
\bfsc{SingleTran} & 68.6 & 74.7 & 70.8 & 65.4 & 75.8 & 61.3 & 69.8 & 74.3 & 69.0 & 73.2 & 51.1 & 68.5 \\
$\Delta_{\text{\bfsc{Baseline}}}$ & 11.5 & 17.4 & 15.6 & 8.0 & 20.4 & 8.7 & 14.9 & 13.1 & 13.8 & 20.2 & 6.9 & 13.7 \\
\midrule
\bfsc{Holdout} & 70.4 & 77.8 & 75.1 & 68.9 & 75.5 & 63.3 & 70.7 & 76.4 & 70.8 & 73.7 & 51.3 & 70.4 \\
$\Delta_{\text{\bfsc{Baseline}}}$ & 13.3 & 20.5 & 19.8 & 11.5 & 20.1 & 10.7 & 15.8 & 15.2 & 15.6 & 20.7 & 7.1 & 15.5 \\
\midrule
\bfsc{AllLangs} & 72.0 & 82.5 & 79.6 & 75.9 & 77.6 & 68.2 & 73.0 & 80.3 & 73.1 & 75.1 & 57.8 & 74.1 \\
$\Delta_{\text{\bfsc{Baseline}}}$ & 14.9 & 25.2 & 24.4 & 18.5 & 22.2 & 15.6 & 18.1 & 19.1 & 17.9 & 22.1 & 13.7 & 19.2 \\
\bottomrule
\multicolumn{13}{c}{\rule{0pt}{2ex}\textbf{Structural Probe Results: Distance Spearman Correlation (DSpr.)}}\\
\toprule
\bfsc{Linear} &.573 &.570 &.533 &.567 &.589 &.489 &.564 &.598 &.578 &.543 &.493 &.554 \\
\bfsc{mBERTRand} &.657 &.658 &.672 &.659 &.693 &.611 &.621 &.710 &.656 &.608 &.590 &.649 \\
\midrule
\bfsc{In-Lang} &.822 &.845 &.846 &.817 &.859 &.813 &.812 &.864 &.807 &.798 &.777 &.824 \\
$\Delta_{\text{\bfsc{Baseline}}}$ &.165 &.187 &.174 &.158 &.166 &.202 &.191 &.154 &.151 &.190 &.187 &.175 \\
\midrule
\bfsc{SingleTran} &.774 &.801 &.807 &.773 &.838 &.732 &.787 &.836 &.772 &.771 &.655 &.777 \\
$\Delta_{\text{\bfsc{Baseline}}}$&.117 &.143 &.135 &.114 &.145 &.121 &.166 &.126 &.117 &.163 &.064 &.128 \\
\midrule
\bfsc{Holdout} &.779 &.821 &.824 &.788 &.838 &.744 &.792 &.840 &.776 &.775 &.664 &.786 \\
$\Delta_{\text{\bfsc{Baseline}}}$&.122 &.163 &.152 &.129 &.146 &.133 &.171 &.130 &.121 &.166 &.074 &.137 \\
\midrule
\bfsc{AllLangs} &.795 &.839 &.836 &.806 &.848 &.777 &.802 &.853 &.789 &.783 &.717 &.804 \\
$\Delta_{\text{\bfsc{Baseline}}}$&.138 &.181 &.165 &.147 &.155 &.165 &.181 &.143 &.134 &.174 &.127 &.156 \\
\bottomrule
\end{tabular}
\caption{Performance (in UUAS and DSpr.) of the structural probe trained on the following cross-lingual sources of data: the evaluation language (\bfsc{In-Lang}); the single other best language (\bfsc{SingleTran}); all other languages (\bfsc{Holdout}); and all languages, including the evaluation language (\bfsc{AllLangs}).  
Note that all improvements over baseline ($\Delta_{\text{\bfsc{Baseline}}}$) are reported against the stronger of our two baselines per-language.
}
\label{tab:results}
\end{table*}

\subsection{Results}

We find that the syntactic probe recovers syntactic trees across all the languages we investigate, achieving on average an improvement of 22 points UUAS and 0.175 DSpr. over both baselines (\tabref{results}, section \textsc{In-Language}).\footnote{Throughout this paper, we report improvement over the stronger of our two baselines per-language.}

Additionally, the probe achieves significantly higher UUAS (on average, 9.3 points better on absolute performance and 6.7 points better on improvement over baseline) on Western European languages.\footnote{
 Here, we define Western European as Czech, English, French, German, and Spanish.
}.  
Such languages have been shown to have better performance on recent shared task results on multilingual parsing \citep[e.g.][]{zeman-etal-2018-conll}.
However, we do not find a large improvement when evaluated on DSpr. (0.041 DSpr. absolute, -0.013 relative).

\begin{figure}
    \centering
    \includegraphics[width=\linewidth]{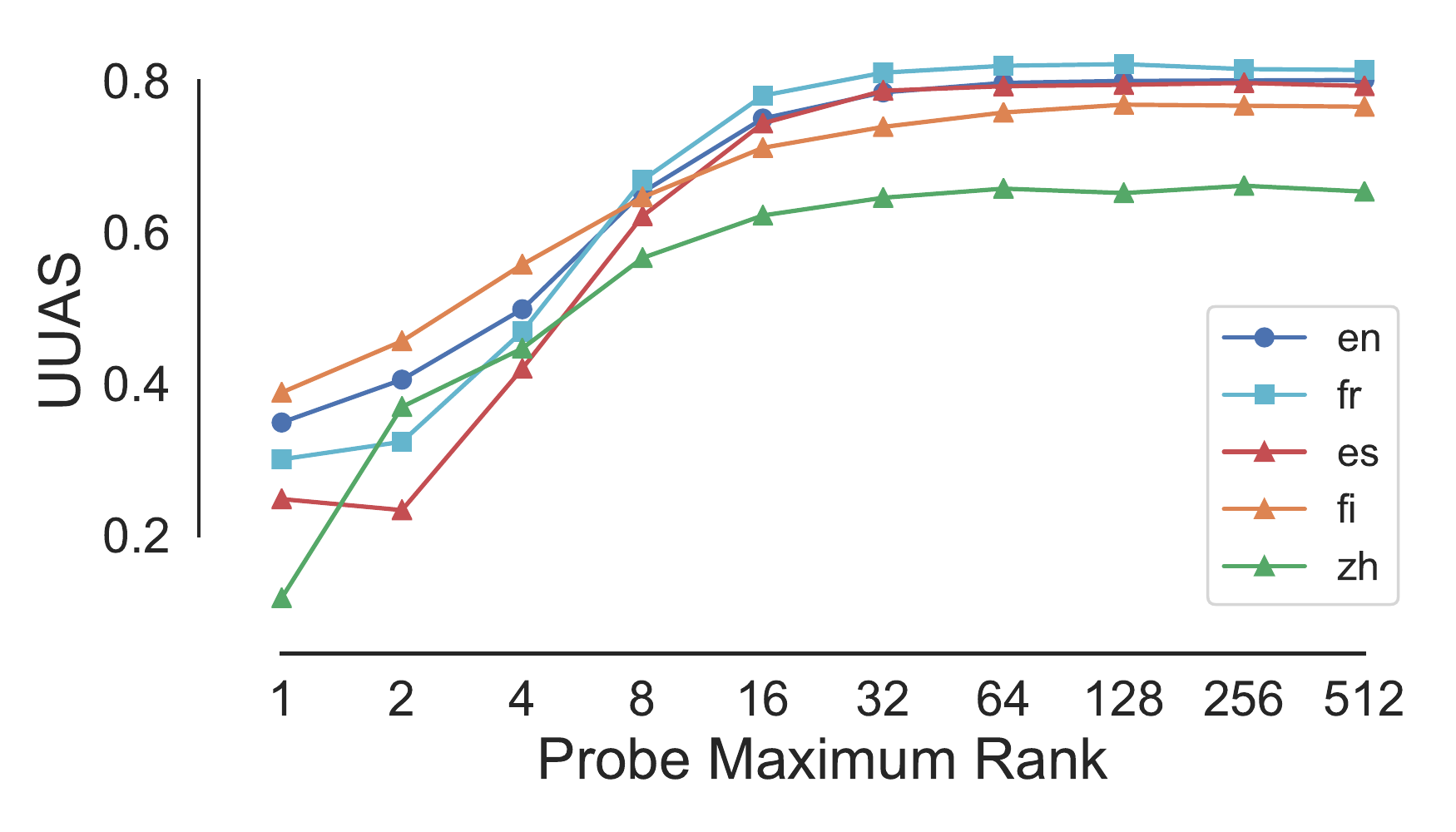}
    \caption{\label{fig:rank}Parse distance tree reconstruction accuracy (UUAS) for selected languages at layer 7 when the linear transformation is constrained to varying maximum dimensionality.}
\end{figure}

\begin{figure}
    \centering
    \includegraphics[width=\linewidth]{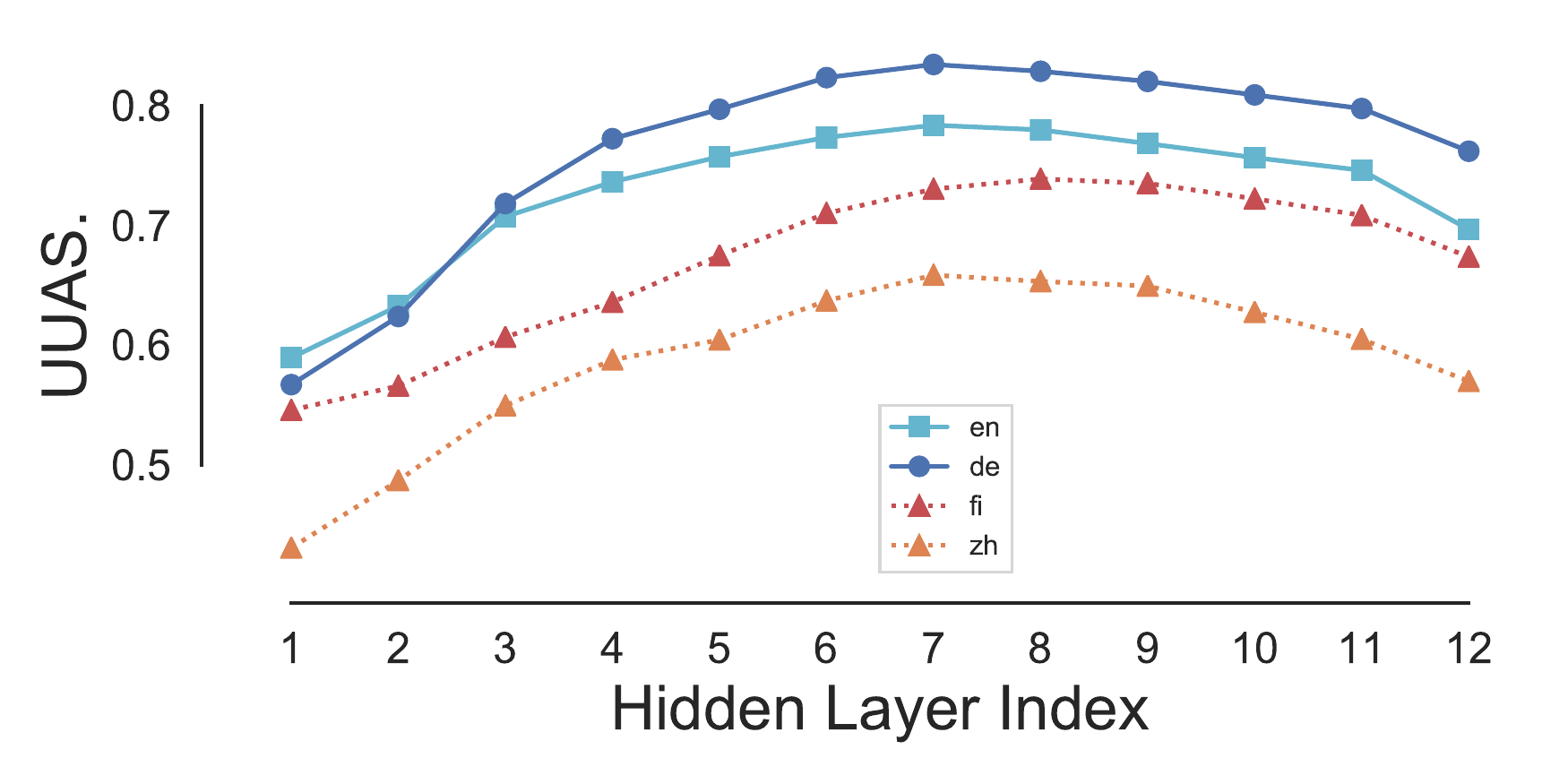}
    \caption{\label{fig:layers}Parse distance tree reconstruction accuracy (UUAS) on layers 1--12 for selected languages, with probe maximum rank 128.}
\end{figure}

We find that across all languages we examine, the structural probe most effectively recovers tree structure from the 7th or 8th mBERT layer (\figref{layers}).
Furthermore, increasing the probe maximum rank beyond approximately 64 or 128 gives no further gains, implying that the syntactic subspace is a small part of the overall mBERT representation, which has dimension 768 (\figref{rank}).

These results closely correspond to the results found by \citet{hewitt2019structural} for an equivalently sized monolingual English model trained and evaluated on the Penn Treebank \cite{marcus1993building}, suggesting that mBERT behaves similarly to monolingual BERT in representing syntax. 

\section{Cross-Lingual Probing}
\label{sec:transfer}

\subsection{Transfer Experiments} 

We now evaluate the extent to which Multilingual BERT's syntactic subspaces are similar across languages.
To do this, we evaluate the performance of a structural probe when evaluated on a language unseen at training time.  
If a probe trained to predict syntax from representations in language $i$ also predicts syntax in language $j$, this is evidence that mBERT's syntactic subspace for language $i$ also encodes syntax in language $j$, and thus that syntax is encoded similarly between the two languages.

Specifically, we evaluate the performance of the structural probe in the following contexts:

\begin{itemize}
    \item \textbf{Direct transfer}, where we train on language $i$ and evaluate on language $j$.
    \item \textbf{Hold-one-out transfer}, where we train on all languages other than $j$ and evaluate on language $j$.
\end{itemize}

\subsection{Joint Syntactic Subspace}

Building off these cross-lingual transfer experiments, we investigate whether there exists a single joint syntactic subspace that encodes syntax in all languages, and if so, the degree to which it does so.  To do so, we train a probe on the concatenation of data from all languages, evaluating it on the concatenation of validation data from all languages.

\subsection{Results}
We find that mBERT's syntactic subspaces are transferable across all of the languages we examine.
Specifically, transfer from the best source language (chosen \textit{post hoc} per-language) achieves on average an improvement of 14 points UUAS and 0.128 DSpr. over the best baseline (\tabref{results}, section \textsc{SingleTran}).\footnote{For full results, consult Appendix Table 1.}
Additionally, our results demonstrate the existence of a cross-lingual syntactic subspace; on average, a holdout subspace trained on all languages but the evaluation language achieves an improvement of 16 points UUAS and 0.137 DSpr. over baseline, while a joint \textsc{AllLangs} subspace trained on a concatenation of data from all source languages achieves an improvement of 19 points UUAS and 0.156 DSpr. (\tabref{results}, section \textsc{Holdout, AllLangs}).

Furthermore, for most languages, syntactic information embedded in the \textit{post hoc} best cross-lingual subspace accounts for 62.3\% of the total possible improvement in UUAS (73.1\% DSpr.) in recovering syntactic trees over the baseline (as represented by in-language supervision).
Holdout transfer represents on average 70.5\% of improvement in UUAS (79\% DSpr.) over the best baseline, while evaluating on a joint syntactic subspace accounts for 88\% of improvement in UUAS (89\% DSpr.).
These results demonstrate the degree to which the cross-lingual syntactic space represents syntax cross-lingually.

\subsection{Subspace Similarity}

Our experiments attempt to evaluate syntactic overlap through zero-shot evaluation of structural probes.  
In an effort to measure more directly the degree to which the syntactic subspaces of mBERT overlap, we calculate the average principal angle\footnote{\url{https://docs.scipy.org/doc/scipy/reference/generated/scipy.linalg.subspace_angles.html}} between the subspaces parametrized by each language we evaluate, to test the hypothesis that syntactic subspaces which are closer in angle have closer syntactic properties (Table \ref{tab:subspace}).

We evaluate this hypothesis by asking whether closer subspaces (as measured by lower average principal angle) correlate with better cross-lingual transfer performance.
For each language $i$, we first compute an ordering of all other languages $j$ by increasing probing transfer performance trained on $j$ and evaluated on $i$.
We then compute the Spearman correlation between this ordering and the ordering given by decreasing subspace angle.
Averaged across all languages, the Spearman correlation is 0.78 with UUAS, and 0.82 with DSpr., showing that transfer probe performance is substantially correlated with subspace similarity.

\subsection{Extrapolation Testing}
To get a finer-grained understanding of how syntax is shared cross-lingually, we aim to understand whether less common syntactic features are embedded in the same cross-lingual space as syntactic features common to all languages.  
To this end, we examine two syntactic relations---prenominal and postnominal adjectives---which appear in some of our languages but not others.
We train syntactic probes to learn a subspace on languages that primarily only use one ordering (i.e. majority class is greater than 95\% of all adjectives), then evaluate their UUAS score solely on adjectives of the other ordering.
Specifically, we evaluate on French, which has a mix (69.8\% prenominal) of both orderings, in the hope that evaluating both orderings in the same language may help correct for biases in pairwise language similarity.
Since the evaluation ordering is out-of-domain for the probe, predicting evaluation-order dependencies successfully suggests that the learned subspace is capable of generalizing between both kinds of adjectives.

We find that for both categories of languages, accuracy does not differ significantly on either prenominal or postnominal adjectives.  Specifically, for both primarily-prenominal and primarily-postnominal training languages, postnominal adjectives score on average approximately 2 points better than prenominal adjectives (\tabref{extrapolation}).

\begin{table}
\small
\centering
\begin{tabular}{rlll}
\toprule
Language & Prenom. & Postnom. & \% data prenom.\\
\midrule
de & 0.932 & 0.900 & 100.0\% \\
zh & 0.801 & 0.826 & 100.0\% \\
lv & 0.752 & 0.811 & 99.7\% \\
en & 0.906 & 0.898 & 99.1\% \\
fi & 0.834 & 0.840 & 98.5\% \\
cz & 0.830 & 0.894 & 95.4\% \\
\midrule
fa & 0.873 & 0.882 & 9.6\% \\
id & 0.891 & 0.893 & 4.9\% \\
ar & 0.834 & 0.870 & 0.1\% \\
\midrule
\textbf{Average pre:} & 0.843 & 0.862 \\
\textbf{Average post:} & 0.866 & 0.881 \\
\bottomrule
\end{tabular}
\caption{\label{tab:extrapolation}Performance of syntactic spaces trained on various languages on recovering prenominal and postnominal French noun--adjective edges.}

\end{table}

\begin{figure}
    \centering
    \includegraphics[width=\linewidth]{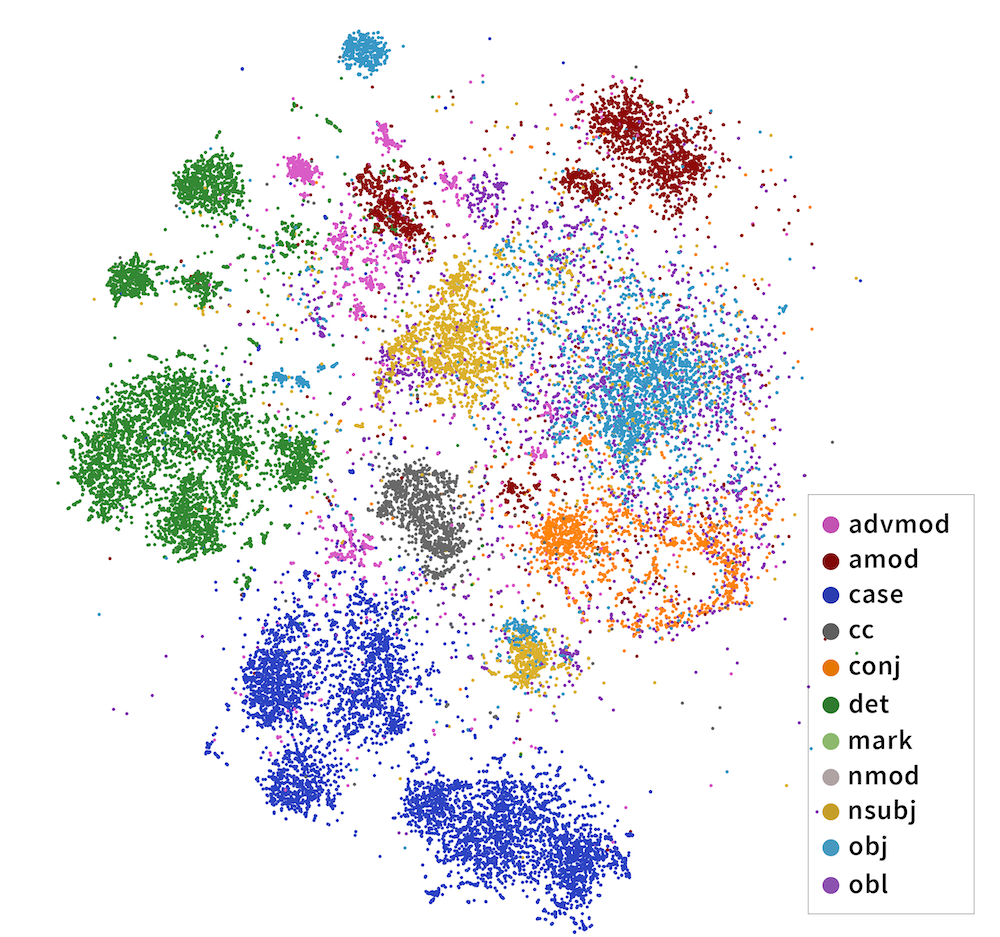}
    \caption{\label{fig:holdout_es}t-SNE visualization of syntactic differences in Spanish projected into a holdout subspace (learned by a probe trained to recover syntax trees in languages other than Spanish).  Despite never seeing a Spanish sentence during probe training, the subspace captures a surprisingly fine-grained view of Spanish dependencies.}
\end{figure}

\section{mBERT Dependency Clusters Capture Universal Grammatical Relations}

\subsection{Methodology}

Given the previous evidence that mBERT shares syntactic representations cross-lingually, we aim to more qualitatively examine the nature of syntactic dependencies in syntactic subspaces.
Let $\mathcal{D}$ be a dataset of parsed sentences, and the linear transformation $B \in \mathbb{R}^{k \times m}$ define a $k$-dimensional syntactic subspace.
For every non-root word and hence syntactic dependency in $\mathcal{D}$ (since every word is a dependent of some other word or an added ROOT symbol), we calculate the $k$-dimensional \textit{head-dependent vector} between the head and the dependent after projection by $B$.
Specifically, for all head-dependent pairs $(w_\text{head}, w_\text{dep})$, we compute $v_\text{diff} = B(\h_\text{head} - \h_\text{dep})$.
We then visualize all differences over all sentences in two dimensions using t-SNE \citep{tsne}.

\subsection{Experiments}

\begin{figure*}[ht]%
    \centering
        \begin{subfigure}{0.8\linewidth}
        \centering
        \includegraphics[width=\linewidth]{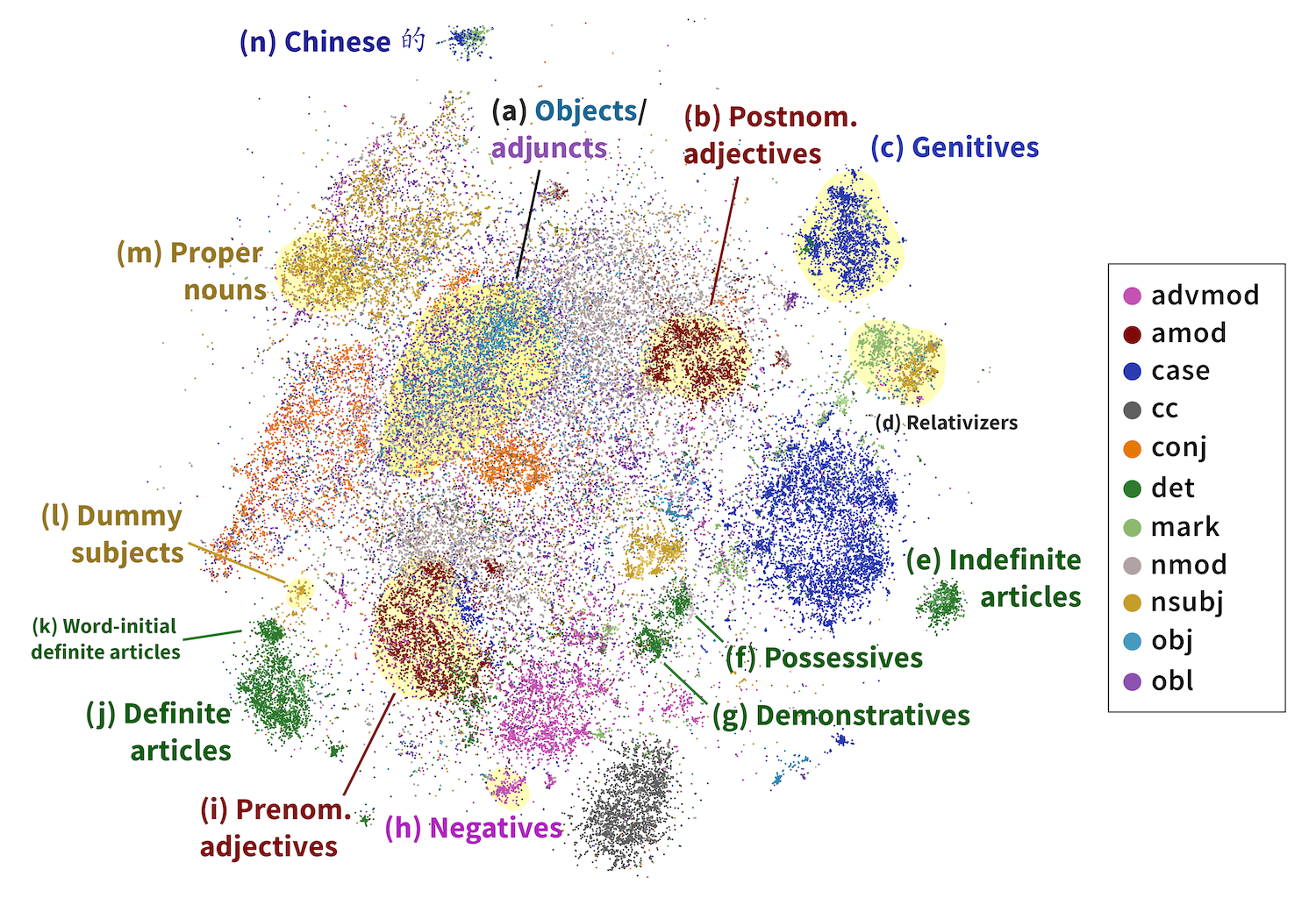}
        \label{subfiga}
        \end{subfigure}\hfill%
        \par 
        \begin{subfigure}{\linewidth}
        \small
            \centering
            \caption*{Example sentences (trimmed for clarity).  Heads in \textbf{bold}; dependents in \bfit{bold italic}.}
\begin{CJK*}{UTF8}{bkai}
            \begin{tabular}{rll}
            \toprule
            \textbf{(b) Postnominal adjectives} & \texttt{fr} & Le gaz développe ses \textbf{applications} \bfit{domestiques}. \\
            & \texttt{id} & \textbf{Film} \bfit{lain} yang menerima penghargaan istimewa. \\ 
            \setfarsi 
            & \texttt{fa} & {\tiny \RL{ وی تصميمات ... اوپک در تنظيم قيمت {\bf نفت \textit{خام}} و ... را ... مؤثر ... دانست }} \\  
            \midrule
            \textbf{(c) Genitives} & \texttt{en} & The assortment \bfit{of} \textbf{customers} adds entertainment. \\
            & \texttt{es} & Con la recuperaci\'{o}n \bfit{de} la \textbf{democracia} y las libertades\\
            & \texttt{lv} & Svešiniece piecēlās, atvadījās \bfit{no} vecā \textbf{vīra} \\
            \midrule
            \textbf{(j) Definite articles} & \texttt{en} & The value of \textbf{\textit{the}} highest \textbf{bid} \\
            & \texttt{fr} & Merak est une ville d'Indon\'{e}sie sur \textbf{\textit{la}} c\^{o}te \textbf{occidentale}. \\
            & \texttt{de} &Selbst mitten in \textbf{\textit{der}} \textbf{Woche} war das Lokal gut besucht. \\
            \bottomrule
            \end{tabular}
                \end{CJK*}
        \label{subfigb}%
        \end{subfigure}\hfill%
        \caption{t-SNE visualization of 100,000 syntactic difference vectors projected into the cross-lingual syntactic subspace of Multilingual BERT.   
        We exclude \rel{punct} and visualize the top 11 dependencies remaining, which are collectively responsible for 79.36\% of the dependencies in our dataset.
        Clusters of interest highlighted in yellow; linguistically interesting clusters labeled. }
    \label{fig:multiling-annotated}
\end{figure*}

As with multilingual probing, one can visualize head-dependent vectors in several ways; we present the following experiments:

\begin{itemize}
    \item dependencies from one language, projected into a different language's space (\figref{enfr_clusters})
    \item dependencies from one language, projected into a holdout syntactic space trained on all other languages (\figref{holdout_es})
    \item dependencies from all languages, projected into a joint syntactic space trained on all languages (\figref{multiling-annotated})
\end{itemize}

For all these experiments, we project into 32-dimensional syntactic spaces.\footnote
  {We reduce the dimensionality of the subspaces here as compared to our previous experiments to match t-SNE suggestions and more aggressively filter non-syntactic information.}
Additionally, we expose a web interface for visualization in our GitHub repository.\footnote{ \url{https://github.com/ethanachi/multilingual-probing-visualization/blob/master/visualization.md}}

\setcode{utf8}

\subsection{Results}

When projected into a syntactic subspace determined by a structural probe, we find that difference vectors separate into clusters reflecting linguistic characteristics of the dependencies.  
The cluster identities largely overlap with (but do not exactly agree with) dependency labels as defined by Universal Dependencies (\figref{multiling-annotated}).  
Additionally, the clusters found by mBERT are highly multilingual. When dependencies from several languages are projected into the same syntactic subspace, whether trained monolingually or cross-lingually, we find that dependencies of the same label share the same cluster (e.g. \figref{enfr_clusters}, which presents both English and French syntactic difference vectors projected into an English subspace).

\subsection{Finer-Grained Analysis} \label{sec_finergrained}

Visualizing syntactic differences in the syntactic space provides a surprisingly nuanced view of the native distinctions made by mBERT.
In \figref{multiling-annotated}, these differences are colored by gold UD dependency labels.    A brief summary is as follows:

    \begin{CJK*}{UTF8}{bkai}
    \paragraph{Adjectives} Universal Dependencies categorizes all adjectival noun modifiers under the \rel{amod} relation.  However, we find that mBERT splits adjectives into two groups: prenominal adjectives in cluster (b) (e.g., Chinese \textbf{獨特的}地理) and postnominal adjectives in cluster (u) (e.g., French \textit{applications \textbf{domestiques}}).
    \end{CJK*}
    \paragraph{Nominal arguments} mBERT maintains the UD distinction between subject (\rel{nsubj}) and object (\rel{obj}).  Indirect objects (\rel{iobj}) cluster with direct objects.  Interestingly, mBERT generally groups adjunct arguments (\rel{obl}) with \rel{nsubj} if near the beginning of a sentence and \rel{obj} otherwise.
    \paragraph{Relative clauses} In the languages in our dataset, there are two major ways of forming relative clauses.  Relative pronouns (e.g., English \textit{the man \textbf{who} is \textup{hungry}} are classed by Universal Dependencies as being an \rel{nsubj} dependent, while subordinating markers (e.g., English \textit{I know \textbf{that} she saw me}) are classed as the dependent of a \rel{mark} relation.  However, mBERT groups both of these relations together, clustering them distinctly from most \rel{nsubj} and \rel{mark} relations.
    \begin{CJK*}{UTF8}{bkai}
    \paragraph{Negatives} Negative adverbial modifiers (English \textit{not}, Farsi \RL{غیر}, Chinese 不) are not clustered with other adverbial syntactic relations (\rel{advmod}), but form their own group (h).\footnote{Stanford Dependencies and Universal Dependencies v1 had a separate \rel{neg} dependency, but it was eliminated in UDv2.}
    \end{CJK*}
    \paragraph{Determiners} The linguistic category of determiners (\rel{det}) is split into definite articles (i), indefinite articles (e), possessives (f), and demonstratives~(g).  Sentence-initial definite articles (k) cluster separately from other definite articles (j). 
    \paragraph{Expletive subjects} Just as in UD, with the separate relation \rel{expl}, expletive subjects, or third-person pronouns with no syntactic meaning (e.g. English \textit{\textbf{It} is cold}, French \textit{\textbf{Il} faudrait}, Indonesian \textit{\textbf{Yang} menjadi masalah kemudian}), cluster separately (k) from other \rel{nsubj} relations (small cluster in the bottom left).

Overall, mBERT draws slightly different distinctions from Universal Dependencies.  Although some are more fine-grained than UD, others appear to be more influenced by word order, separating relations that most linguists would group together.  Still others are valid linguistic distinctions not distinguished by the UD standard.

\subsection{Discussion}

Previous work has found that it is possible to recover dependency labels from mBERT embeddings, in the form of very high accuracy on dependency label probes \cite{liu2019linguistic,tenney2019what}.  
However, although we know that dependency label probes are able to use supervision to map from mBERT's representations to UD dependency labels, this does not provide full insight into the nature of (or existence of) latent dependency label structure in mBERT.
By contrast, in the structural probe, $B$ is optimized such that $\|v_\text{diff}\|_2 \approx 1$, but no supervision as to dependency label is given.
The contribution of our method is thus to provide a view into mBERT's ``own'' dependency label representation.
In Appendix~\ref{sec_additional_viz}, Figure~\ref{fig:mbert_rand_t-SNE}, we provide a similar visualization as applied to \textsc{mBERTrand}, finding much less cluster coherence.

\subsection{Probing as a window into representations}
Our head-dependent vector visualization uses a supervised probe, but its objects of study are properties of the representation \textit{other} than those relating to the probe supervision signal.  
Because the probe never sees supervision on the task we visualize for, the visualized behavior cannot be the result of the probe memorizing the task, a problem in probing methodology \cite{hewitt2019control}.  
Instead, it is an example of using probe supervision to focus in on aspects that may be drowned out in the original representation. 
However, the probe's linear transformation may not pick up on aspects that are of causal influence to the model.

\section{Related Work}

\paragraph{Cross-lingual embedding alignment}
\citet{lample2018word} find that independently trained monolingual word embedding spaces in ELMo are isometric under rotation.  
Similarly, \citet{schuster2019cross} and \citet{wang2019crosslingual} geometrically align contextualized word embeddings trained independently.
\citet{wu2019emerging} find that cross-lingual transfer in mBERT is possible even without shared vocabulary tokens, which they attribute to this isometricity.
In concurrent work, \citet{cao2020multilingual} demonstrate that mBERT embeddings of similar words in similar sentences across languages are approximately aligned already, suggesting that mBERT also aligns semantics across languages.
\citet{karthikeyan2020crosslingual} demonstrate that strong cross-lingual transfer is possible without any word piece overlap at all.

\paragraph{Analysis with the structural probe}
In a monolingual study, \citet{reif2019visualizing} also use the structural probe of \citet{hewitt2019structural} as a tool for understanding the syntax of BERT.
They plot the words of individual sentences in a 2-dimensional PCA projection of the structural probe distances, for a geometric visualization of individual syntax trees.
Further, they find that distances in the mBERT space separate clusters of word senses for the same word type.

\paragraph{Understanding representations}
\citet{pires2019multilingual} find that cross-lingual BERT representations share a common subspace representing useful linguistic information. 
\citet{libovicky2019language} find that mBERT representations are composed of a language-specific component and a language-neutral component.  Both \citet{libovicky2019language} and \citet{kudugunta2019investigating} perform SVCCA on LM representations extracted from mBERT and a massively multilingual transformer-based NMT model, finding language family-like clusters.  

\citet{li2019specializing} present a study in syntactically motivated dimensionality reduction; they find that after being passed through an information bottleneck and dimensionality reduction via t-SNE, ELMo representations cluster naturally by UD part of speech tags. 
Unlike our syntactic dimensionality reduction process, the information bottleneck is directly supervised on POS tags, whereas our process receives no linguistic supervision other than unlabeled tree structure.  
In addition, the reduction process, a feed-forward neural network, is more complex than our linear transformation.

\citet{singh2019bert} evaluate the similarity of mBERT representations using Canonical Correlation Analysis (CCA), finding that overlap among subword tokens accounts for much of the representational similarity of mBERT\@.  
However, they analyze cross-lingual overlap across all components of the mBERT representation, whereas we evaluate solely the overlap of syntactic subspaces.
Since syntactic subspaces are at most a small part of the total BERT space, 
these are not necessarily mutually contradictory with our results. 
In concurrent work, \citet{julian2020asking} also extend probing methodology, extracting latent ontologies from contextual representations without direct supervision.

\section{Discussion}

Language models trained on large amounts of text have been shown to develop surprising emergent properties; of particular interest is the emergence of non-trivial, easily accessible linguistic properties seemingly far removed from the training objective.
For example, it would be a reasonable strategy for mBERT to share little representation space between languages, effectively learning a private model for each language and avoiding destructive interference.
Instead, our transfer experiments provide evidence that at a syntactic level, mBERT shares portions of its representation space between languages.
Perhaps more surprisingly, we find evidence for fine-grained, cross-lingual syntactic distinctions in these representations.
Even though our method for identifying these distinctions lacks dependency label supervision, we still identify that mBERT has a cross-linguistic clustering of grammatical relations that qualitatively overlaps considerably with the Universal Dependencies formalism.

\paragraph{The UUAS metric}
We note that the UUAS metric alone is insufficient for evaluating the accuracy of the structural probe.
While the probe is optimized to directly recreate parse distances, (that is, $d_B(\mathbf{h}_i^\ell, \mathbf{h}_j^\ell) \approx d_T^\ell(w_i^\ell, w_j^\ell)$) a perfect UUAS score under the minimum spanning tree construction can be achieved by ensuring that $d_B(\mathbf{h}_i^\ell, \mathbf{h}_j^\ell)$ is small if there is an edge between $w_i^\ell$ and $w_j^\ell$, and large otherwise, instead of accurately recreating distances between words connected by longer paths.
By evaluating Spearman correlation between all pairs of words, one directly evaluates the extent to which the ordering of words $j$ by distance to each word $i$ is correctly predicted, a key notion of the geometric interpretation of the structural probe.
See \citet{maudslay2020tale} for further discussion.

\paragraph{Limitations} Our methods are unable to tease apart, for all pairs of languages, whether transfer performance is caused by subword overlap \cite{singh2019bert} or by a more fundamental sharing of parameters, though we do note that language pairs with minimal subword overlap do exhibit non-zero transfer, both in our experiments and in others \cite{karthikeyan2020crosslingual}.
Moreover, while we quantitatively evaluate cross-lingual transfer in recovering dependency distances, we only conduct a qualitative study in the unsupervised emergence of dependency labels via t-SNE. Future work could extend this analysis to include quantitative results on the extent of agreement with UD.  We acknowledge as well issues in interpreting t-SNE plots \cite{wattenberg2016use}, and include multiple plots with various hyperparameter settings to hedge against this confounder in Figure~\ref{fig:differing_ppl}.   

Future work should explore other multilingual models like XLM and XLM-RoBERTa \cite{lample2019cross} and attempt to come to an understanding of the extent to which the properties we've discovered have causal implications for the decisions made by the model, a claim our methods cannot support.

\section{Acknowledgements}
We would like to thank Erik Jones, Sebastian Schuster, and Chris Donahue for helpful feedback and suggestions.  We would also like to thank the anonymous reviewers and area chair Adina Williams for their helpful comments on our draft.

\clearpage
\bibliography{acl2019} 
\bibliographystyle{acl_natbib}

\appendix

\section{Additional Syntactic Difference Visualizations} \label{sec_additional_viz}

\subsection{Visualization of All Relations}

\begin{figure}[ht]
    \centering
    \includegraphics[width=\linewidth]{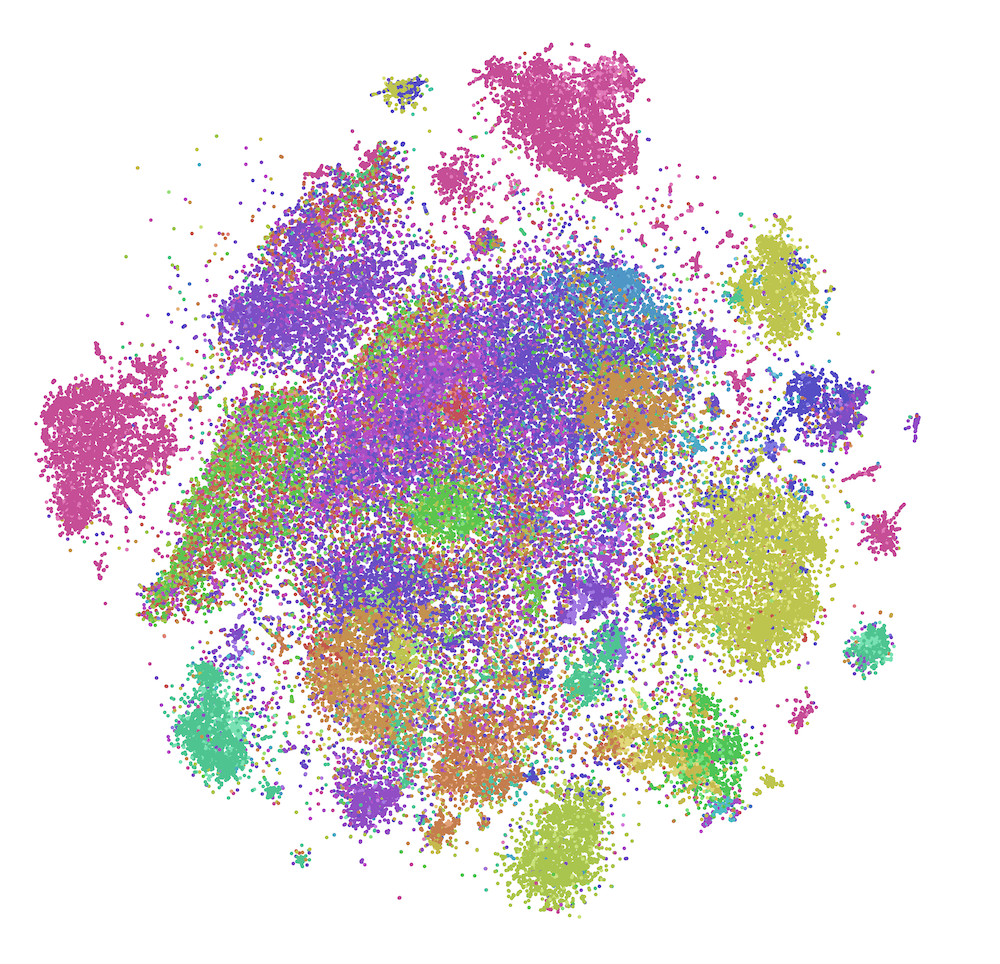}
    \caption{t-SNE visualization of dependency head-dependent pairs projected into the cross-lingual syntactic subspace of Multilingual BERT\@.  Colors correspond to gold UD dependency type labels, which are unlabeled given that there are 43 in this visualization.}
    \label{fig:all}
\end{figure}

In our t-SNE visualization of syntactic difference vectors projected into the cross-lingual syntactic subspace of Multilingual BERT (\figref{multiling-annotated}), we only visualize the top 11 relations, excluding \rel{punct}.  This represents 79.36\% of the dependencies in our dataset.  In \figref{all}, we visualize all 36 relations in the dataset.

\subsection{Visualization with Randomly-Initialized Baseline}

\begin{figure}[ht]
    \centering
    \includegraphics[width=\linewidth]{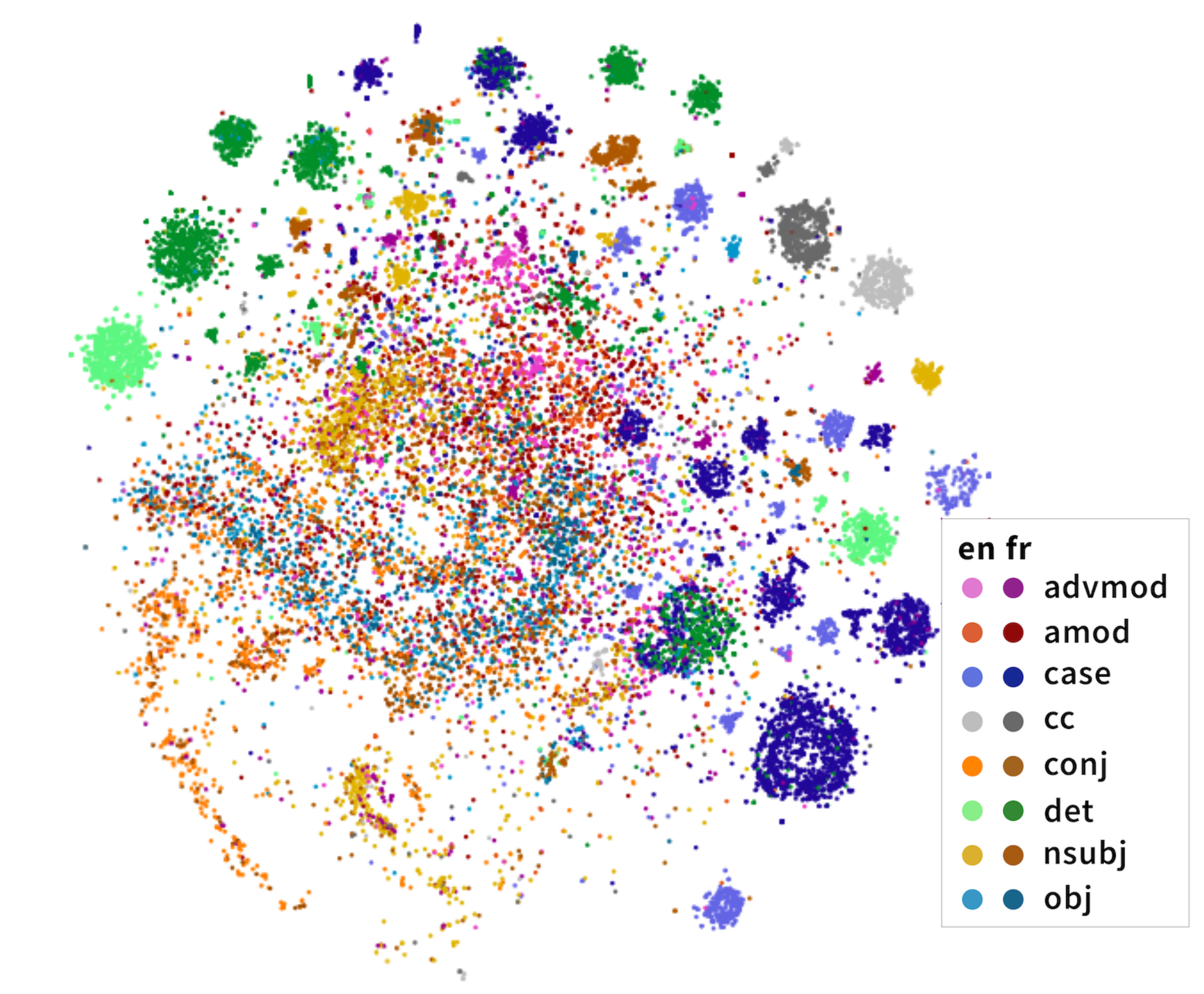}
    \caption{\label{fig:mbert_rand_t-SNE}t-SNE visualization of head-dependent dependency pairs belonging to selected dependencies in English and French, projected into a syntactic subspace of \textsc{mBERTrand}, as learned on English syntax trees. Colors correspond to gold UD dependency type labels.
}
\end{figure}

In \figref{mbert_rand_t-SNE}, we present a visualization akin to \figref{enfr_clusters}; however, both the head-dependency representations, as well as the syntactic subspace, are derived from \textsc{mBERTRand}. 
Clusters around the edges of the figure are primarily type-based (e.g. one cluster for the word \textit{for} and another for \textit{pour}), and there is insignificant overlap between clusters with parallel syntactic functions from different languages.

\section{Alternative Dimensionality Reduction Strategies}

In an effort to confirm the level of clarity of the clusters of dependency types which emerge from syntactic difference vectors, we examine simpler strategies for dimensionality reduction.

\subsection{PCA for Visualization Reduction}

We project difference vectors as previously into a 32-dimensional syntactic subspace.  However, we visualize in 2 dimensions using PCA instead of t-SNE.  There are no significant trends evident.

\begin{figure}[ht]
    \centering
    \includegraphics[width=0.8\linewidth,keepaspectratio]{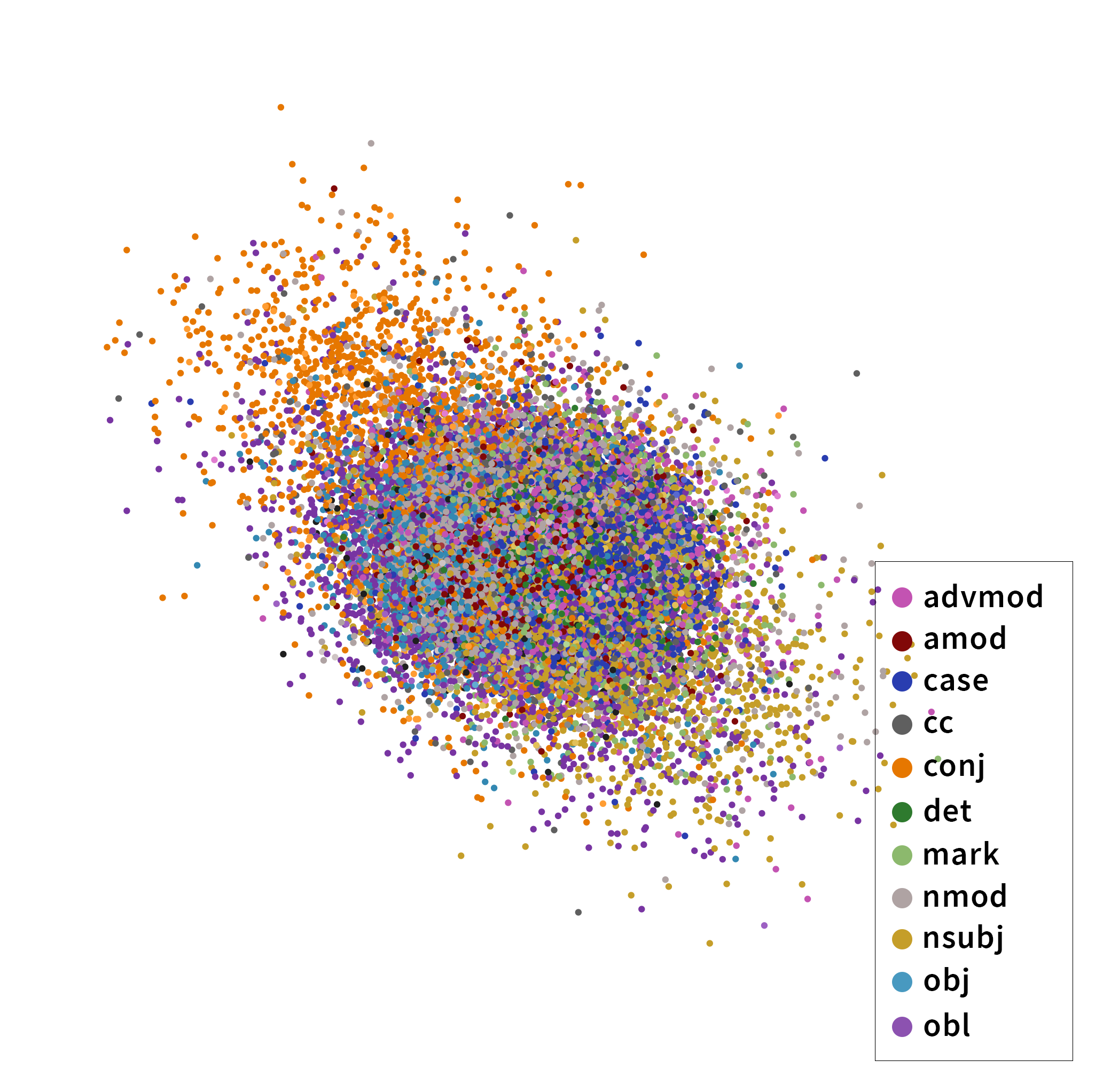}
    \caption{\label{fig:multilingclusters-pca}Syntactic difference vectors visualized after dimensionality reduction with PCA, instead of t-SNE, colored by UD dependency types.  There are no significant trends evident.}
\end{figure}

\subsection{PCA for Dimensionality Reduction}

Instead of projecting difference vectors into our syntactic subspace, we first reduce them to a 32-dimensional representation using PCA,\footnote
  {This is of equal dimensionality to our syntactic subspace.} 
then reduce to 2 dimensions using t-SNE as previously.  

\begin{figure}[ht]
    \centering
    \includegraphics[height=0.8\linewidth,keepaspectratio]{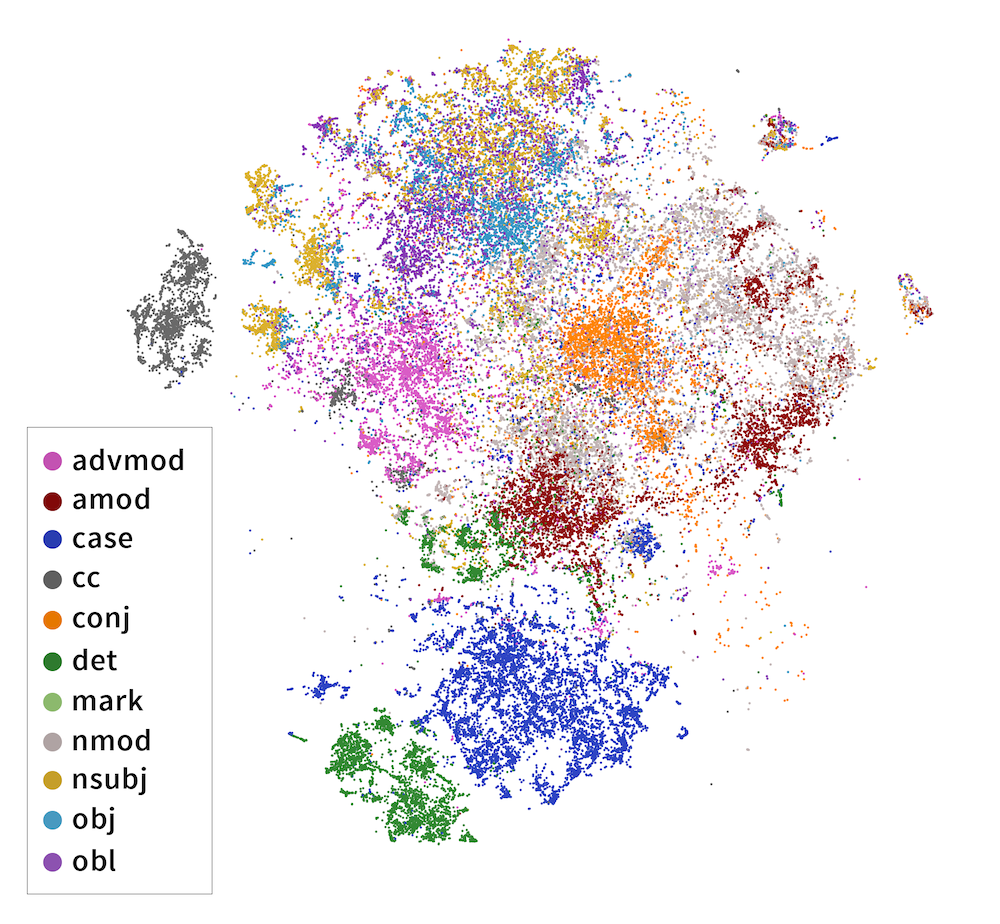}
    \caption{\label{fig:pca_red}t-SNE visualization of syntactic differences in all languages we study, projected to 32 dimensions using PCA.}
\end{figure}

We find that projected under PCA, syntactic difference vectors still cluster into major groups, and major trends are still evident (\figref{pca_red}).  
In addition, many finer-grained distinctions are still apparent (e.g. the division between common nouns and pronouns).
However, in some cases, the clusters are motivated less by syntax and more by semantics or language identities.  For example:

\begin{itemize}
    \item The \rel{nsubj} and \rel{obj} clusters overlap, unlike our syntactically-projected visualization, where there is clearer separation.
    \item Postnominal adjectives, which form a single coherent cluster under our original visualization scheme, are split into several different clusters, each primarily composed of words from one specific language.
    \item There are several small monolingual clusters without any common syntactic meaning, mainly composed of languages parsed more poorly by BERT (i.e. Chinese, Arabic, Farsi, Indonesian).
\end{itemize}

\begin{table*}[t]
\scriptsize
\centering
\begin{tabular}{llllllllllllllll}
\multicolumn{16}{c}{\textbf{Structural Probe Results: Undirected Unlabeled Attachment Score (UUAS)}}\\ 
\toprule
Tgt \textbackslash Src & ar & cz & de & en & es & fa & fi & fr & id & lv & zh & linear & rand & holdout & all \\
\midrule
ar & 72.7 & 68.6 & 66.6 & 65.3 & 67.5 & 64.0 & 60.8 & 68.1 & 65.3 & 60.1 & 53.4 & 57.1 & 49.8 & 70.4 & 72.0 \\
cz & 57.5* & 83.6 & 74.7 & 72.6 & 71.1 & 63.5 & 68.9 & 71.5 & 62.4 & 71.0 & 58.0 & 45.4 & 57.3 & 77.8 & 82.5 \\
de & 49.3 & 70.2 & 83.5 & 70.8 & 68.2 & 58.7 & 61.1 & 70.6 & 56.9* & 62.0 & 52.0* & 42.8 & 55.2 & 75.1 & 79.6 \\
en & 47.2 & 61.2 & 65.0 & 79.8 & 63.9 & 50.8 & 55.3 & 65.4 & 54.5 & 54.0 & 50.5 & 41.5 & 57.4 & 68.9 & 75.9 \\
es & 52.0 & 67.2 & 69.8 & 69.4 & 79.7 & 56.9 & 56.8 & 75.8 & 61.0 & 55.6 & 49.2 & 44.6 & 55.3 & 75.5 & 77.6 \\
fa & 51.7 & 61.3 & 60.3 & 57.0 & 57.8 & 70.8 & 53.7 & 59.7 & 56.5 & 53.1 & 49.7 & 52.6 & 43.2 & 63.3 & 68.2 \\
fi & 55.5 & 69.8 & 68.4 & 66.6 & 66.0 & 60.2 & 76.5 & 66.0 & 61.2 & 68.2 & 59.2 & 50.1 & 54.9 & 70.7 & 73.0 \\
fr & 50.8* & 67.8 & 73.0 & 70.0 & 74.3 & 56.9 & 55.9 & 84.0 & 60.9 & 55.1 & 49.6 & 46.4 & 61.2 & 76.4 & 80.3 \\
id & 57.1 & 66.3 & 67.4 & 63.6 & 67.0 & 61.0 & 59.2 & 69.0 & 74.8 & 57.5 & 54.6 & 55.2 & 53.2 & 70.8 & 73.1 \\
lv & 56.9* & 73.2 & 69.2 & 69.1 & 67.0 & 61.5 & 70.8 & 66.7 & 61.1 & 77.0 & 60.7 & 47.0 & 53.0 & 73.7 & 75.1 \\
zh & 41.2* & 49.7 & 49.6 & 51.1 & 47.3 & 42.7* & 48.1 & 47.9 & 44.5* & 47.2 & 65.7 & 44.2 & 41.1 & 51.3 & 57.8 \\
\bottomrule 
\\
\multicolumn{16}{c}{\rule{0pt}{2ex}\textbf{Structural Probe Results: Distance Spearman Correlation (DSpr.)}} \\ 
\toprule
Tgt \textbackslash Src & ar & cz & de & en & es & fa & fi & fr & id & lv & zh & linear & rand & holdout & all \\
\midrule
ar & .822 & .772 & .746 & .744 & .774 & .730 & .723 & .770 & .750 & .722 & .640 & .573 & .657 & .779 & .795 \\
cz & .730 & .845 & .799 & .781 & .801 & .741 & .782 & .796 & .745 & .791 & .656 & .570 & .658 & .821 & .839 \\
de & .690 & .807 & .846 & .792 & .792 & .736 & .767 & .796 & .723 & .765 & .652* & .533 & .672 & .824 & .836 \\
en & .687 & .765 & .764 & .817 & .770 & .696 & .732 & .773 & .720 & .725 & .655 & .567 & .659 & .788 & .806 \\
es & .745 & .821 & .812 & .806 & .859 & .741 & .775 & .838 & .777 & .774 & .669 & .589 & .693 & .838 & .848 \\
fa & .661 & .732 & .724 & .706 & .705 & .813 & .683 & .714 & .686 & .684 & .629 & .489 & .611 & .744 & .777 \\
fi & .682* & .787 & .771 & .756 & .764 & .712 & .812 & .762 & .715 & .781 & .658 & .564 & .621 & .792 & .802 \\
fr & .731* & .810 & .816 & .806 & .836 & .738 & .767 & .864 & .776 & .760 & .674 & .598 & .710 & .840 & .853 \\
id & .715 & .757 & .752 & .739 & .765 & .718 & .714 & .772 & .807 & .704 & .657 & .578 & .656 & .776 & .789 \\
lv & .681 & .771 & .746 & .737 & .745 & .699 & .763 & .740 & .698 & .798 & .644 & .543 & .608 & .775 & .783 \\
zh & .538* & .655 & .644 & .644 & .633 & .593* & .652 & .638 & .584* & .639 & .777 & .493 & .590 & .664 & .717 \\
\bottomrule
\end{tabular}
\caption{Performance (in UUAS and DSpr.) on transfer between all language pairs in our dataset. All runs were repeated 3 times; runs for which the range in performance exceeded 2 points (for UUAS) or 0.02 (for DSpr.) are marked with an asterisk (*).}
\label{tab:transfer}
\end{table*}

\begin{table*}[t]
\small
\centering
\begin{tabular}{rllllllllllll}
\toprule
 & ar & cz & de & en & es & fa & fi & fr & id & lv & zh \\
\midrule 
ar & 0.000 & 1.044 & 1.048 & 1.049 & 1.015 & 1.046 & 1.058 & 1.022 & 1.031 & 1.059 & 1.076 \\
cz & 1.044 & 0.000 & 0.982 & 1.017 & 0.970 & 1.064 & 1.021 & 1.007 & 1.053 & 1.011 & 1.083 \\
de & 1.048 & 0.982 & 0.000 & 1.005 & 0.973 & 1.044 & 1.017 & 0.971 & 1.022 & 1.029 & 1.065 \\
en & 1.049 & 1.017 & 1.005 & 0.000 & 0.983 & 1.051 & 1.033 & 0.994 & 1.035 & 1.040 & 1.060 \\
es & 1.015 & 0.970 & 0.973 & 0.983 & 0.000 & 1.038 & 1.023 & 0.936 & 1.010 & 1.024 & 1.065 \\
fa & 1.046 & 1.064 & 1.044 & 1.051 & 1.038 & 0.000 & 1.060 & 1.028 & 1.040 & 1.063 & 1.069 \\
fi & 1.058 & 1.021 & 1.017 & 1.033 & 1.023 & 1.060 & 0.000 & 1.020 & 1.042 & 1.011 & 1.058 \\
fr & 1.022 & 1.007 & 0.971 & 0.994 & 0.936 & 1.028 & 1.020 & 0.000 & 0.993 & 1.028 & 1.041 \\
id & 1.031 & 1.053 & 1.022 & 1.035 & 1.010 & 1.040 & 1.042 & 0.993 & 0.000 & 1.051 & 1.052 \\
lv & 1.059 & 1.011 & 1.029 & 1.040 & 1.024 & 1.063 & 1.011 & 1.028 & 1.051 & 0.000 & 1.068 \\
zh & 1.076 & 1.083 & 1.065 & 1.060 & 1.065 & 1.069 & 1.058 & 1.041 & 1.052 & 1.068 & 0.000 \\
\bottomrule
\end{tabular}
\caption{\label{tab:subspace}Subspace angle overlap as evaluated by the pairwise mean principal angle between subspaces}

\end{table*}

\begin{table*}[t]
\small
\centering
\begin{tabular}{rlllllllllllll}
\toprule
\bf Language & ar & cz & de & en & es & fa & fi & fr & id & lv & zh \\
\midrule
\bf Spearman Correl. (UUAS) & 0.88 & 0.85 & 0.87 & 0.91 & 0.91 & 0.48 & 0.85 & 0.89 & 0.71 & 0.90 & 0.41 \\
\bf Spearman Correl. (DSpr.) & 0.95 & 0.96 & 0.95 & 0.96 & 0.97 & 0.50 & 0.90 & 0.93 & 0.72 & 0.94 & 0.23\\
\bottomrule
\end{tabular}
\caption{\label{tab:spearman} The Spearman correlation between two orderings of all languages for each language $i$. The first ordering of languages is given by (negative) subspace angle between the $B$ matrix of language $i$ and that of all languages. The second ordering is given by the structural probe transfer accuracy from all languages (including $i$) to $i$. This is repeated for each of the two structural probe evaluation metrics.}
\end{table*}

\section{Additional Experiment Settings}

\subsection{Pairwise Transfer}

We present full pairwise transfer results in Table \ref{tab:transfer}.  
Each experiment was run 3 times with different random seeds; experiment settings with range in UUAS greater than 2 points are labeled with an asterisk (*). 

\subsection{Subspace Overlap}

Table \ref{tab:subspace} presents the average principal angle between the subspaces parametrized by each language we evaluate. Table \ref{tab:spearman} contains the per-language Spearman correlation between the ordering given by (negative) subspace angle and structural probe transfer accuracy, reported both on UUAS and DSpr.

\section{Data Sources}

We use the following UD corpora in our experiments: Arabic-PADT, Chinese-GSD, Czech-PDT, English-EWT, Finnish-TDT, French-GSD, German-GSD, Indonesian-GSD, Latvian-LVTB, Persian-Seraji, and Spanish-Ancora.

\section{t-SNE reproducibility}

\begin{figure*}
\centering
        \begin{subfigure}{0.24\linewidth}
        \includegraphics[width=1\linewidth]{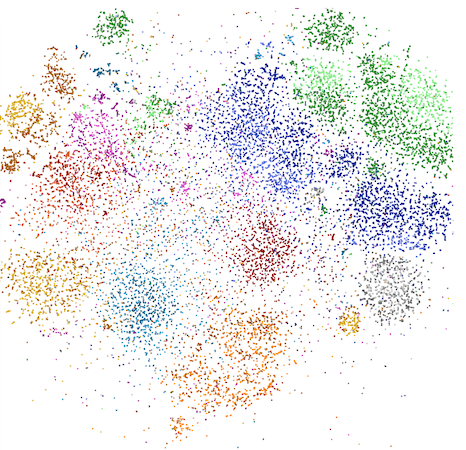}
        \end{subfigure}
        \begin{subfigure}{0.24\linewidth}
        \includegraphics[width=1\linewidth]{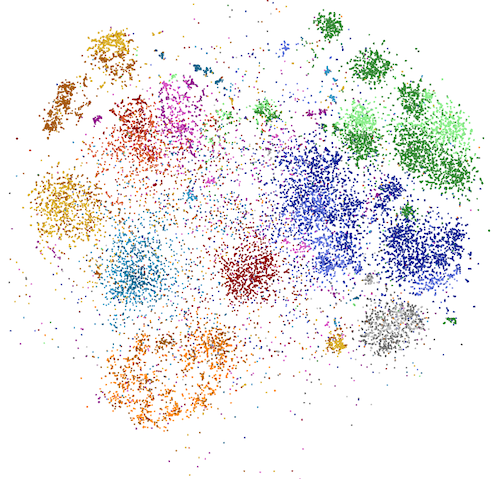}
        \end{subfigure}
        \begin{subfigure}{0.24\linewidth}
        \includegraphics[width=1\linewidth]{en-fr_final.png}
        \end{subfigure}
        \begin{subfigure}{0.24\linewidth}
        \includegraphics[width=1\linewidth]{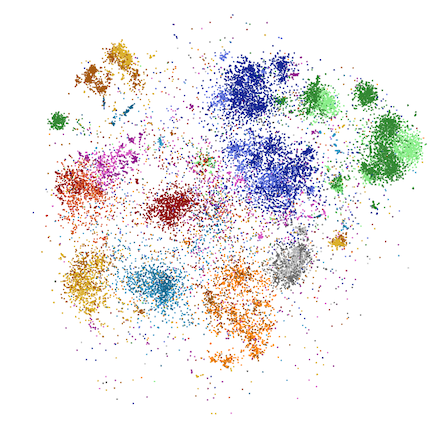}
        \end{subfigure}
    \caption{\label{fig:differing_ppl}t-SNE visualization of head-dependent dependency pairs belonging to selected dependencies in English and French, projected into a syntactic subspace of Multilingual BERT, as learned on English syntax trees. Colors correspond to gold UD dependency type labels, as in Figure~\ref{fig:enfr_clusters}, varying the perplexity (PPL) t-SNE hyperparmeter. From left to right, figures correspond to PPL $5,10, 30, 50$, spanning the range of PPL suggested by \citet{tsne}. Cross-lingual dependency label clusters are exhibited across all four figures. } 
\end{figure*}

Previous work \cite{wattenberg2016use} has investigated issues in the interpretability of tSNE plots.  Given the qualitative nature of our experiments, to avoid this confounder, we include multiple plots with various settings of the perplexity hyperparameter in \figref{differing_ppl}.

\end{document}


\maketitle

\section{Appendix: Additional Syntactic Difference Visualizations}

\subsection{Visualization of All Relations}

\begin{figure}[h]
    \centering
    \includegraphics[width=\linewidth]{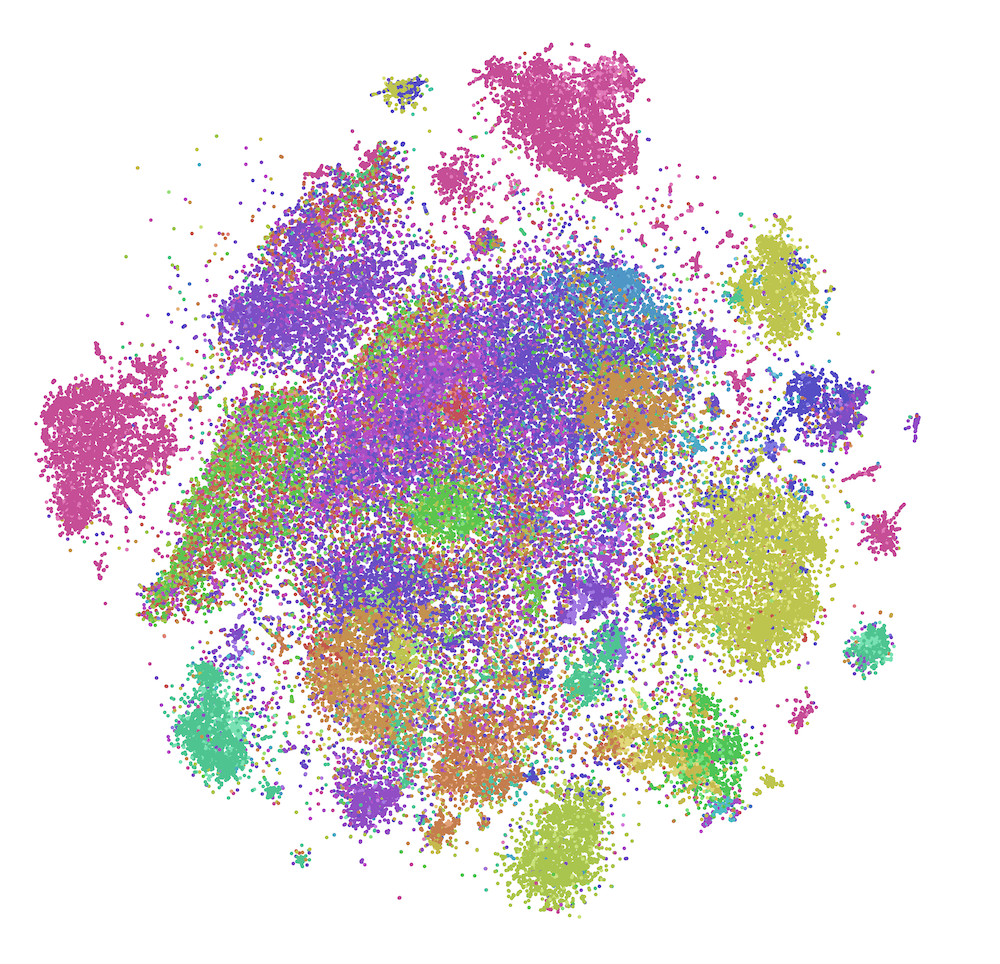}
    \caption{tSNE visualization of head-dependent pairs dependencies
    projected into the cross-lingual syntactic subspace of Multilingual BERT\@.  Colors correspond to gold UD dependency type labels. Although neither mBERT nor our probe is ever trained on UD dependency labels, the clusters in the learned representation substantially capture human analyses of dependency types.}
    \label{fig:all}
\end{figure}

In our main visualization, we only visualize the top 11 relations, excluding \rel{punct}.  Below, we visualize all 36 relations in the dataset (\figref{all}).

\begin{table*}[t]
\small
\centering
\begin{tabular}{llllllllllllllll}
\toprule
Dest \textbackslash Src & ar & cz & de & en & es & fa & fi & fr & id & lv & zh & linear & rand & holdout & all \\
\midrule
ar  & 72.8 & 68.4 & 67.2 & 65.2 & 67.4 & 63.4 & 60.8 & 67.0 & 65.1 & 59.6 & 53.8 & 57.1 & 50.0 & 70.4 & 72.0 \\
cz  & 59.1 & 83.7 & 74.9 & 71.9 & 70.8 & 63.2 & 69.0 & 69.8 & 62.1 & 70.9 & 57.2 & 45.4 & 57.3 & 77.8 & 82.5 \\
de  & 49.0 & 70.3 & 83.4 & 70.4 & 68.1 & 59.3 & 62.4 & 67.8 & 58.1 & 61.7 & 52.8 & 42.8 & 55.3 & 75.1 & 79.6 \\
en  & 47.4 & 61.3 & 65.3 & 80.1 & 62.4 & 51.1 & 53.8 & 63.7 & 55.3 & 55.1 & 50.1 & 41.5 & 57.6 & 68.9 & 75.9 \\
es  & 52.8 & 67.4 & 69.5 & 69.1 & 79.4 & 56.0 & 56.9 & 73.6 & 60.9 & 55.9 & 49.1 & 44.6 & 55.4 & 75.5 & 77.6 \\
fa  & 51.5 & 61.5 & 60.6 & 57.7 & 57.9 & 70.7 & 54.1 & 57.6 & 57.0 & 52.9 & 48.4 & 52.6 & 42.6 & 63.3 & 68.2 \\
fi  & 57.2 & 69.8 & 68.4 & 66.5 & 65.9 & 60.2 & 76.3 & 65.1 & 60.6 & 68.3 & 58.9 & 50.1 & 54.7 & 70.7 & 73.0 \\
fr  & 48.0 & 67.9 & 72.9 & 70.1 & 73.5 & 56.6 & 55.9 & 81.3 & 60.9 & 55.0 & 49.7 & 46.4 & 61.0 & 76.4 & 80.3 \\
id  & 56.8 & 66.6 & 67.2 & 64.1 & 66.9 & 61.1 & 58.7 & 67.5 & 74.4 & 58.1 & 53.6 & 55.2 & 53.2 & 70.8 & 73.1 \\
lv  & 57.9 & 73.3 & 69.6 & 68.8 & 67.0 & 61.1 & 70.9 & 66.2 & 61.2 & 77.1 & 59.9 & 47.0 & 53.2 & 73.7 & 75.1 \\
zh  & 42.1 & 49.1 & 49.9 & 50.5 & 46.9 & 44.1 & 48.6 & 46.9 & 46.0 & 46.5 & 66.3 & 44.2 & 40.9 & 51.3 & 57.8 \\
\bottomrule
\end{tabular}
\caption{Full cross-lingual probe results}
\label{tab:transfer}
\end{table*}

\section{Appendix: Alternative Dimensionality Reduction Strategies}

In an effort to confirm the level of clarity of the clusters of dependency types which emerge from syntactic difference vectors, we examine simpler strategies for dimensionality reduction.

\subsection{PCA for Dimensionality Reduction}

Here, instead of projecting difference vectors into our syntactic subspace, we first reduce them to a 32-dimensional representation using PCA,\footnote
  {This is of equal dimensionality to our syntactic subspace.} 
then reduce to 2 dimensions using tSNE as previously.  

We find that projected under PCA, syntactic difference vectors still cluster into major groups, and major trends are still evident (Figure \figref{pca_red}).  
In addition, many finer-grained distinctions are still apparent (e.g. division between common nouns and pronouns).
However, in some cases, the clusters are motivated less by syntax and more by semantics or language identities.  For example:
\begin{itemize}
    \item The \rel{nsubj} and \rel{obj} clusters overlap, unlike our syntactically-projected visualization, where there is clearer separation.
    \item Postnominal adjectives, which form a single coherent cluster under our original visualization scheme, are split into several different clusters, each of which mainly has words from one specific language.
    \item There are several small monolingual clusters without any common syntactic meaning, mainly composed of languages parsed more poorly by BERT (i.e. Chinese, Arabic, Farsi, Indonesian).
\end{itemize}

\subsection{PCA for Visualization Reduction}

Here, we project difference vectors as previously into a 32-dimensional syntactic subspace.  However, we visualize in 2 dimensions using PCA instead of tSNE.  There are no significant trends evident.

\section{Appendix: Detailed Transfer Probe Performance}

We present pairwise transfer results in Table \ref{tab:transfer}.


\begin{table*}[t]
\small
\centering
\begin{tabular}{rllllllllllll}
\toprule
 & ar & cz & de & en & es & fa & fi & fr & id & lv & zh \\
\midrule 
ar & 0.000 & 1.044 & 1.048 & 1.049 & 1.015 & 1.046 & 1.058 & 1.022 & 1.031 & 1.059 & 1.076 \\
cz & 1.044 & 0.000 & 0.982 & 1.017 & 0.970 & 1.064 & 1.021 & 1.007 & 1.053 & 1.011 & 1.083 \\
de & 1.048 & 0.982 & 0.000 & 1.005 & 0.973 & 1.044 & 1.017 & 0.971 & 1.022 & 1.029 & 1.065 \\
en & 1.049 & 1.017 & 1.005 & 0.000 & 0.983 & 1.051 & 1.033 & 0.994 & 1.035 & 1.040 & 1.060 \\
es & 1.015 & 0.970 & 0.973 & 0.983 & 0.000 & 1.038 & 1.023 & 0.936 & 1.010 & 1.024 & 1.065 \\
fa & 1.046 & 1.064 & 1.044 & 1.051 & 1.038 & 0.000 & 1.060 & 1.028 & 1.040 & 1.063 & 1.069 \\
fi & 1.058 & 1.021 & 1.017 & 1.033 & 1.023 & 1.060 & 0.000 & 1.020 & 1.042 & 1.011 & 1.058 \\
fr & 1.022 & 1.007 & 0.971 & 0.994 & 0.936 & 1.028 & 1.020 & 0.000 & 0.993 & 1.028 & 1.041 \\
id & 1.031 & 1.053 & 1.022 & 1.035 & 1.010 & 1.040 & 1.042 & 0.993 & 0.000 & 1.051 & 1.052 \\
lv & 1.059 & 1.011 & 1.029 & 1.040 & 1.024 & 1.063 & 1.011 & 1.028 & 1.051 & 0.000 & 1.068 \\
zh & 1.076 & 1.083 & 1.065 & 1.060 & 1.065 & 1.069 & 1.058 & 1.041 & 1.052 & 1.068 & 0.000 \\
\bottomrule
\end{tabular}
\caption{Subspace angle overlap as evaluated by the pairwise mean principal angle between subspaces}
\label{tab:subspace}
\end{table*}

\begin{table*}[t]
\small
\centering
\begin{tabular}{rlllllllllllll}
\toprule
\bf Language & ar & cz & de & en & es & fa & fi & fr & id & lv & zh \\
\midrule
\bf Correlation & 0.882 & 0.855 & 0.873 & 0.909 & 0.909 & 0.482 & 0.852 & 0.893 & 0.709 & 0.897 & 0.409 \\
\bottomrule
\end{tabular}
\caption{Spearman Correlation}
\label{tab:spearman}
\end{table*}

\clearpage

\begin{figure}[ht]
    \centering
    \includegraphics[height=0.8\linewidth,keepaspectratio]{pca_red.png}
    \caption{tSNE visualization of syntactic differences in all languages we study, projected to 32 dimensions using PCA.}
    \label{fig:pca_red}
\end{figure}

\newpage

\begin{figure}[ht]
    \centering
    \includegraphics[height=0.8\linewidth,keepaspectratio]{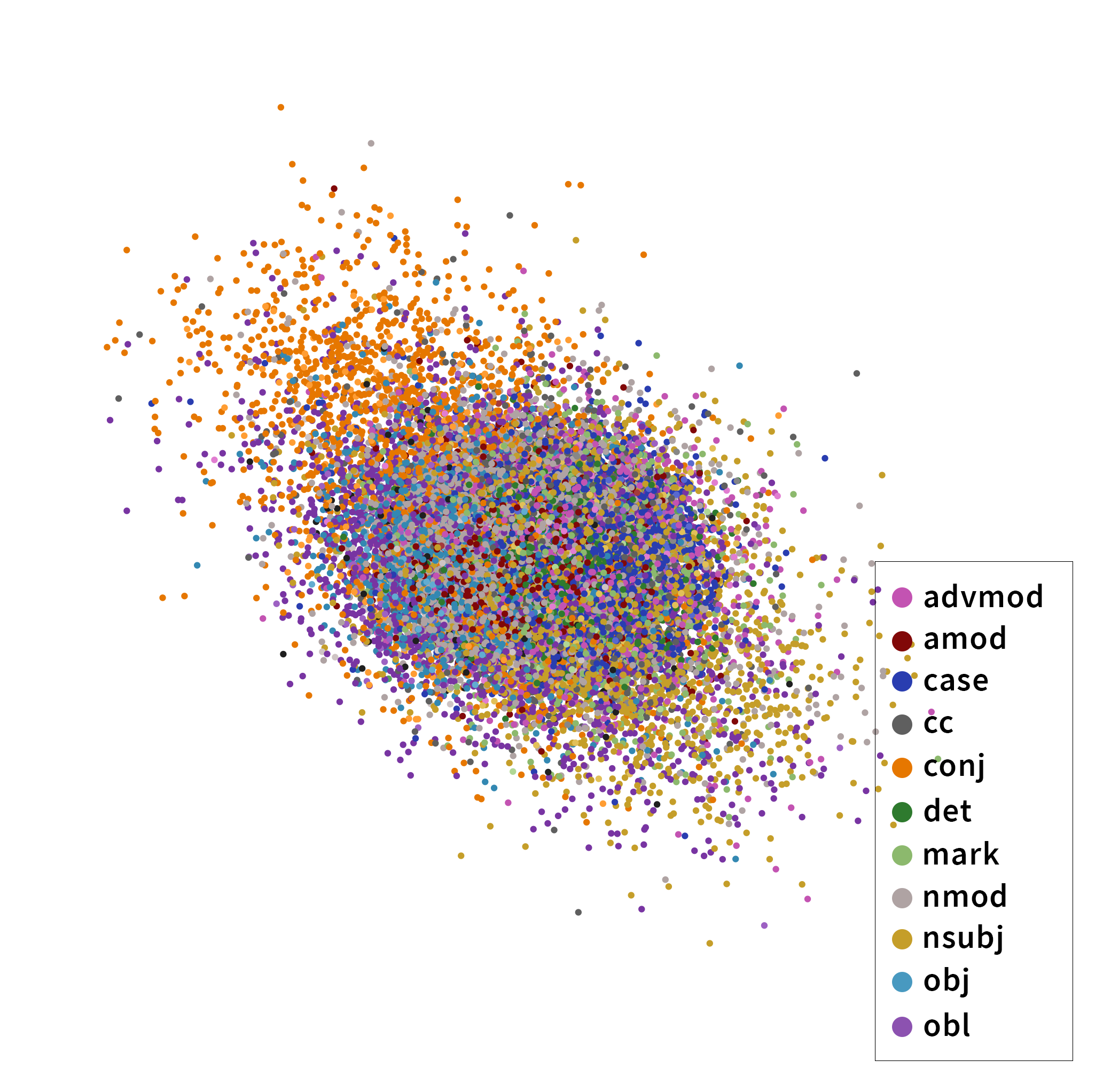}
    \caption{Syntactic difference vectors visualized after dimensionality reduction with PCA, instead of tSNE, colored by UD dependency types.  There are no significant trends evident.}
    \label{fig:multilingclusters-pca}
\end{figure}